%% file: article-ch-pfia.tex
\documentclass[english]{pfia}

\usepackage{tikz}
\usepackage{float} 
\usepackage{standalone}
\usetikzlibrary{trees,shapes.geometric, arrows.meta, positioning,fit,backgrounds, shadows, babel, decorations.pathreplacing, patterns, calc}
\usepackage{tabularx}   
\usepackage{multirow}   
\usepackage{makecell}
\usepackage{xcolor}
\usepackage{amsmath}
\usepackage{amssymb}
\usepackage{graphicx} 
\usepackage{booktabs}
\usepackage{authblk}

\definecolor{afiablue}{RGB}{61,159,207}
\definecolor{afiared}{RGB}{167,75,68}
\definecolor{afialightblue}{RGB}{158,193,232}

\title{\textbf{SeqGPT: A Constrained Transformer Agent for the Inverse Design of Multi-Panel Composite Structures}}

\author{Driss Chraibi\textsuperscript{1}, Alejandro García Pis\textsuperscript{1}, Stéphane Grihon\textsuperscript{1}, Sixin Zhang\textsuperscript{2}}

\affil{\textsuperscript{1}Airbus Operations SAS, Toulouse, France \\ 
       \textsuperscript{2}IRIT, Université de Toulouse, INP-Toulouse}

\begin{document}

\maketitle

\begin{resume}
La conception inverse de structures composites exige de réconcilier des cibles de performance continues avec un espace de design discret et contraint. C’est une approche classique lorsque des stratégies d’optimisation numérique bi-niveaux sont employées. Ce problème inverse combinatoire devient particulièrement critique dans les structures multi-panneaux en raison du blending—une contrainte de continuité ou compatibilité  globale entre les différents empilements. Ce travail introduit SeqGPT, un agent Transformer conditionnel conçu pour remplacer les méthodes itératives, coûteuses en temps de calcul. Afin de garantir la continuité globale et la validité manufacturière par construction, nous employons une stratégie de décodage neuro-symbolique hybride. À chaque pli, le modèle prédit une distribution conditionnelle qui guide une Constrained Beam Search, dans laquelle toute branche violant les règles de blending est strictement élaguée. Les expériences numériques sur le benchmark du fer à cheval à 18 panneaux démontrent que SeqGPT génère des solutions quasi-instantanément avec des performances en flambement comparables aux méthodes évolutionnaires, offrant une accélération considérable par rapport à l’état de l’art.
\end{resume}

\begin{motscles}
IA Neuro-symbolique, Transformer, Modèle conditionnel, Optimisation, Composites.
\end{motscles}

\begin{abstract}
Optimizing composite stacking sequences to match continuous targets (e.g., Lamination or Buckling Parameters) with discrete manufacturing constraints represents a challenging combinatorial inverse problem that regularly occurs in composite design especially when numerical optimization approaches are used (bi-step, bi-level configurations). In multi-panel configurations, this complexity is further intensified by blending—a global compatibility/continuity requirement between the different panel stackings. This study presents SeqGPT, a conditional Transformer agent developed to replace computationally expensive iterative methods. To ensure both global continuity and manufacturing feasibility by construction, we implemented a hybrid neuro-symbolic decoding strategy. SeqGPT predicts a conditional distribution that guides a Constrained Beam Search, where any branch violating blending rules is strictly pruned. Numerical experiments on the 18-panel horseshoe benchmark demonstrate that SeqGPT generates solutions near-instantaneously with buckling performance comparable to evolutionary methods, offering a significant speed-up compared to the state of the art.
\end{abstract}

\begin{keywords}
Neuro-Symbolic AI, Transformer, Conditional Model, Optimization, Composite Structures, Constrained Generation.
\end{keywords}

\section{Introduction}

The contemporary aerospace industry is undergoing a progressive substitution of traditional metal alloys by Carbon Fiber Reinforced Polymer (CFRP) composites—which now account for approximately 53\% of the Airbus A350 XWB structural weight. Unlike isotropic metals, composite structures rely on the superposition of unidirectional plies to resist multi-axial forces, allowing for elastic tailoring. However, physically manufacturing such components restricts the design space to a discrete set of standard orientations $\{0^{\circ}, 90^{\circ}, \pm 45^{\circ}\}$. Furthermore, valid solutions must satisfy strict manufacturing ``hard constraints'' (such as symmetry, balance, and contiguity) \cite{irisarri2014}. Matching continuous mechanical targets within this discrete, highly constrained space creates a non-convex optimization problem with exponential complexity.

This complexity is severely amplified when scaling up to large ``one-shot'' structures divided into multiple adjacent panels. In these configurations, the sequence of a single panel is strictly conditioned by the geometric continuity with its neighbors. This requirement, known as \textit{blending}, turns the optimization into a massively coupled problem where decisions on one panel propagate constraints to the entire structure, often leading to prohibitive convergence times \cite{scardaoni2022}.

State-of-the-art methods generally adopt a two-step optimization strategy: first optimizing continuous structural parameters (e.g., lamination or buckling parameters), and then retrieving a discrete stacking sequence that matches these targets \cite{herencia2007}. In contrast, Irisarri et al. \cite{irisarri2014} proposed a monolithic formulation in which blending constraints are enforced directly through the discrete optimization of Stacking Sequence Tables (SST) using a Genetic Algorithm. While effective, this approach suffers from a strong combinatorial growth of the design space. To restore computational efficiency while guaranteeing blending, the Search Propagation Direction (SPD) method was introduced \cite{scardaoni2022}, enforcing continuity by construction through iterative ply drops or insertions. However, the retrieval of admissible stacking sequences within this framework still relies on iterative stochastic solvers, which remains a bottleneck for real-time interactive design.

Recently, deep generative models have emerged as a promising alternative to stochastic search. Notable examples include the Physics-Informed CGAN by Qiu et al. \cite{qiu2022} and LayupFormer introduced by Xie et al. \cite{xie2026}. While these state-of-the-art methods successfully map mechanical targets to valid stacking sequences, they operate at the single-laminate level, producing one layup for a given structure without considering inter-laminate blending constraints. These models are trained for a specific optimization problem, and extending them to new configurations typically requires costly surrogate retraining.

In this work, we share the same intuition as \cite{xie2026}, reformulating stacking sequence retrieval as a conditional generation task. Building on the general solver paradigm enabled by the bi-step optimization strategy and the Search Propagation Direction (SPD) method \cite{scardaoni2022}, we developed SeqGPT, a Transformer-based agent trained to satisfy manufacturing rules by conditioning predictions on non-dimensional buckling parameters. While SPD ensures computational efficiency and blending compliance at a global scale, our hybrid neuro-symbolic approach, inspired by NeuroLogic Decoding \cite{lu2021}, enforces strict global continuity. In this framework, SeqGPT provides probabilistic guidance conditioned on the target structural parameters, while a coupled Constrained Beam Search strictly prunes invalid branches to satisfy blending rules by construction.

The remainder of this paper is organized as follows. Section 2 details the state-of-the-art methods. Section 3 describes the SeqGPT architecture and the neuro-symbolic decoding logic. Finally, Section 4 evaluates the performance of the proposed method against traditional stochastic solvers.
\section{State of the Art: From Composite Optimization to Generative Inference}

\subsection{The Bi-Step Strategy: Bridging Global Sizing and Local Retrieval}
\label{sec:bi_level}

\begin{figure}[htbp]
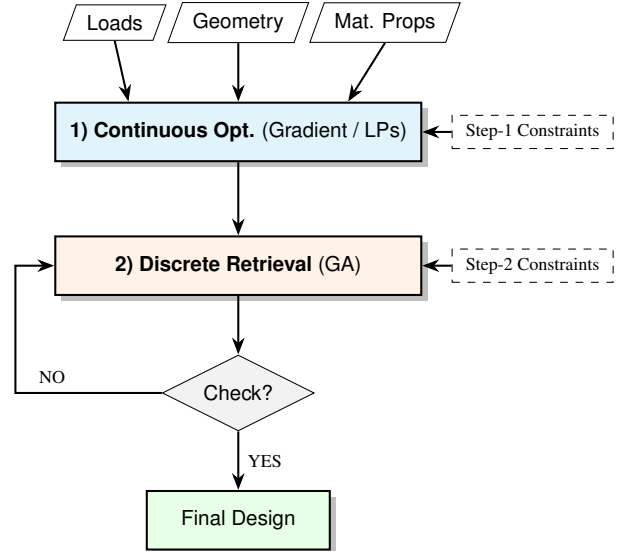

    \centering
    \includestandalone[width=0.45\textwidth]{figures/fig_workflow}
    \caption{The standard Bi-Step Optimization workflow. Step 1 uses fast gradients on the global continuous assembly. Step 2 retrieves the discrete sequences locally, a process that suffers from combinatorial complexity and fails to guarantee global blending.}
    \label{fig:workflow}
\end{figure}

State-of-the-art methods generally adopt a two-step strategy (Fig.~\ref{fig:workflow}) to handle the combinatorial complexity of composite design \cite{herencia2007}.

The first step idealizes the entire multi-panel structure as a continuous domain. At this stage, the solver evaluates the complete structural assembly—incorporating global geometry, boundary conditions, and internal load redistributions—to minimize total mass. Instead of discrete plies, the design variables are relaxed into continuous quantities \cite{herencia2007}. 

Specifically, the solver identifies a global optimum vector $\boldsymbol{\xi}^*$ for each individual panel. This target vector comprises a continuous target thickness $N$ alongside four non-dimensional Nemeth parameters ($\alpha, \beta, \gamma, \delta$). Originally formulated in NASA technical studies \cite{nemeth1985}, these parameters provide a natural abstraction of the laminate's physics. They condense the complex anisotropic bending stiffness matrix into four dimensionless ratios that strictly govern buckling stability: $\alpha$ and $\beta$ characterize the orthotropic bending and torsional stiffness, while $\gamma$ and $\delta$ quantify the destabilizing anisotropic bending-twisting coupling. Because this relaxed feasible domain is convex, gradient-based algorithms can efficiently compute these continuous targets \cite{herencia2007}.

The second step, \textbf{Stacking Sequence Retrieval (SSR)}, constitutes the main computational bottleneck. At this stage, the structure must be translated back into discrete ply angles $\theta \in \Theta = \{0^\circ, \pm 45^\circ, 90^\circ\}$. 

In the standard bi-step approach, SSR is formulated as a purely local inverse problem for each panel:
\begin{equation}
\hat{S} = \arg\min_{S \in \mathcal{X}} \| \Phi(S) - \boldsymbol{\xi}^* \|^2,
\end{equation}
where $\mathcal{X} \subset \Theta^{N}$ is defined as the discrete design space of valid stacking sequences. The mapping $\Phi(S)$ represents the Classical Laminate Theory (CLT) equations, which compute the macroscopic mechanical properties from a discrete sequence of oriented plies $S$. To mitigate unwanted anisotropic effects and ensure structural integrity, the discrete design space $\mathcal{X}$ is bounded by strict manufacturing rules.

\begin{table}[htbp]

\centering

\footnotesize

\caption{Selected subset of critical Manufacturing Constraints (MCs) adapted from \cite{irisarri2014}, defining the feasible design space $\mathcal{X}$.}

\label{tab:manufacturing_constraints}

\renewcommand{\arraystretch}{1.3}

\begin{tabular}{|l|l|p{5.5cm}|}

\hline

\textbf{Scope} & \textbf{ID} & \textbf{Requirement and Technical Description} \\ \hline

\multirow{4}{*}{\textbf{Local}} 

 & \textbf{MC1} & \textbf{Symmetry:} Stacking sequences must be mirrored with respect to the mid-plane, ensuring that the coupling matrix $B$ is zero and decoupling in-plane and bending responses. \\ \cline{2-3}

 & \textbf{MC2} & \textbf{Contiguity:} Limits the number of consecutive plies with identical orientation (typically $\leq 4$) to mitigate the risk of matrix cracking and large-scale delamination. \\ \cline{2-3}

 & \textbf{MC3} & \textbf{Balance:} Requires an equal number of $+45^\circ$ and $-45^\circ$ plies within the laminate to eliminate shear--extension coupling and ensure membrane stability. \\ \cline{2-3}

 & \textbf{MC4} & \textbf{Disorientation:} Limits the orientation change between adjacent plies ($0^\circ/90^\circ$ is not allowed) to reduce interlaminar shear stresses and improve fatigue life. \\ \cline{2-3}

 & \textbf{MC5} & \textbf{Proportion:} Requires each principal ply direction ($0^\circ, 90^\circ, \pm 45^\circ$) to constitute between a minimum (typically 10\%) and a maximum (typically 66\%) of the total laminate to guarantee sufficient multi-axial strength and damage tolerance. \\ \hline \hline

\textbf{Global} 

 & \textbf{MC0} & \textbf{Blending:} Each ply present in a thinner panel must continue into adjacent thicker panels to ensure a continuous load path across the structure. \\ \hline

\end{tabular}

\end{table}

As detailed in Table~\ref{tab:manufacturing_constraints}, these defining rules are partitioned into two categories: \textbf{local constraints} (MC1--MC5) ensuring the structural integrity of each isolated laminate, and a \textbf{global constraint} (MC0) enforcing ply continuity (blending) between adjacent panels.

While the first step efficiently resolves the global mechanics, the SSR step introduces a severe combinatorial challenge. Because the mapping $\Phi$ is non-injective (i.e., multiple stacking sequences can produce the same continuous targets), the search landscape is highly non-convex and filled with local optima \cite{scardaoni2022}. Crucially, formulating SSR as a set of independent local problems strictly confines the search space to the local rules (MC1--MC5). By definition, this panel-by-panel approach cannot enforce the inter-laminate global compatibility constraint (MC0).

\subsection{Enforcing Global Constraints: Blending and the SPD Strategy}

\begin{figure}[htbp]
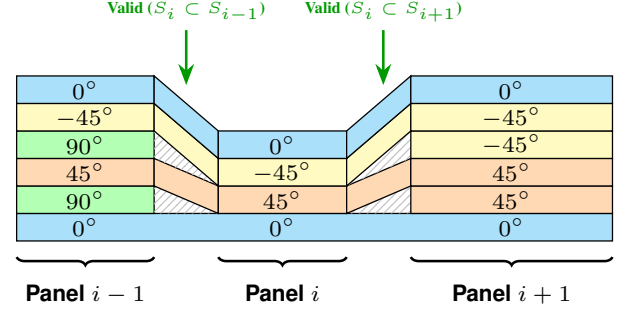

    \centering
    \includestandalone[width=0.45\textwidth]{figures/fig_blending_cross_section}
    \caption{Cross-sectional view of the Blending Constraint. The intermediate panel $S_i$ acts as a topological bridge ($S_i \subset S_{i-1}$ and $S_i \subset S_{i+1}$), ensuring global continuity between neighbors $S_{i-1}$ and $S_{i+1}$ that may otherwise possess mutually incompatible sequences.}
    \label{fig:blending_cross_section}
\end{figure}

To enforce the global blending constraint (MC0, Fig.~\ref{fig:blending_cross_section}), optimization must account for the strict topological dependencies between adjacent panels. Rather than solving isolated problems, each laminate's stacking sequence becomes tightly conditioned by its neighbors. As highlighted by Scardaoni and Montemurro, this non-trivial coupling transforms the sequence retrieval into a massively coupled Constraint Satisfaction Problem (CSP) scaled to the entire structure \cite{scardaoni2022}.

To address this global coupling, Irisarri et al. \cite{irisarri2014} proposed the \textbf{Stacking Sequence Table (SST)}. This method optimizes a single master sequence from which all panels are derived by truncation. While guaranteeing blending by definition, it forces the optimizer (typically a Genetic Algorithm) to evaluate the entire table against the whole structure at every iteration, leading to prohibitive computational costs.

\begin{figure}[htbp]
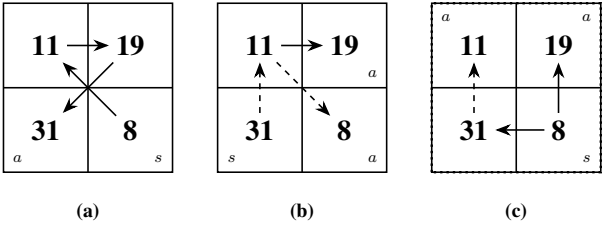

    \centering
    \includestandalone[width=0.45\textwidth]{figures/figspd}
    \caption{Illustration of sequence propagation mechanisms. The numbers within the grids indicate the sequence length $N$. Initial sequences are marked with $s$ and end sequences with $a$. Solid arrows ($\rightarrow$) denote ply insertion, while dashed arrows ($\dashrightarrow$) denote ply drop. (a) Valid propagation utilizing only insertions. (b) Valid propagation mixing insertion and drop with a different start sequence. (c) Invalid cyclic propagation.}
    \label{fig:fig_spd}
\end{figure}

To reduce this global complexity, we adopt the \textit{Search Propagation Direction} (SPD) strategy \cite{scardaoni2022}. SPD decomposes the structure into a directed path of local problems (Fig.~\ref{fig:fig_spd}). In this framework, adjacent panels $S_p$ and $S_q$ (comprising $N_p$ and $N_q$ plies, respectively) are coupled via Propagation Maps $\pi_{p \to q}$:
\begin{itemize}
    \item \textbf{A ply drop ($N_p > N_q$)} uses the map $\pi_{p\to q}$ to select $N_q$ plies to retain from $S_p$ while preserving their relative order.
    
    \item \textbf{A ply insertion ($N_p < N_q$)} ensures that all plies from $S_p$ are inherited, and additional plies are inserted into $S_q$ via a strictly increasing injective map of indices.
\end{itemize}

By fixing this propagation order, the global blending constraint is enforced \textit{by construction}: once the parent sequence $S_p$ is defined, all child sequences $S_q$ along the path are automatically valid. The remaining task is simply to assign the free ply orientations at the root and along the propagation path.

Crucially, the chosen propagation path must not leave physically adjacent panels disconnected. As shown in Fig.~\ref{fig:fig_spd}, paths (a) and (b) are valid because they systematically enforce compatibility rules across physical boundaries. In contrast, path (c) is invalid: the tree branches out, leaving adjacent panels (e.g., the 11-ply and 19-ply laminates) without a direct parent-to-child relationship. Without this explicit evaluation across their shared boundary, the generative model cannot guarantee their final compatibility \cite{scardaoni2022}.

\subsection{From Stochastic Search to Generative Inference}

Although the SPD method simplifies the global topology into a sequence of local problems, assigning the free variables typically relies on iterative stochastic solvers. To overcome this computational latency, recent works have proposed replacing heuristic searches with Deep Generative Models.

\textbf{Conditional GANs (CGAN)} \cite{qiu2022} were proposed as a Physics-Informed approach to generate sequences from mechanical targets. However, as noted by Xie et al. \cite{xie2026}, their stochastic nature lacks interpretability. Furthermore, adversarial training suffers from inherent instabilities and is prone to \textit{mode collapse}, leading to a severe loss of design diversity when scaling to larger sequences.

\textbf{Transformer-Based Architectures} were concurrently investigated by Xie et al. \cite{xie2026}, who introduced \textit{LayupFormer} to formulate sequence retrieval as a linguistic translation task. While promising, their implementation relies on explicit grammar masking applied \textit{after} the Softmax layer and a complex hybrid loss function. Computing this loss requires training and maintaining a high-fidelity differentiable surrogate, introducing significant architectural overhead and complex hyperparameter tuning. Crucially, these generative models operate solely at the single-laminate level, ignoring inter-laminate blending constraints.

\textbf{Our proposed neuro-symbolic approach}, in contrast, explicitly decouples physical performance from topological constraints. The neural agent, SeqGPT, approximates the inverse distribution $P(S \mid \boldsymbol{\xi}^*)$ to inherently satisfy local constraints (MC1--MC5) without intermediate surrogates. To address the global topology, our Constrained Beam Search practically implements the SPD strategy to enforce blending (MC0). By treating inter-panel derivation as a constrained tree search, symbolic logic strictly prunes invalid ply drops and insertions. Throughout this process, SeqGPT acts as a physics-driven heuristic to evaluate and rank the valid branches, transforming an intractable combinatorial problem into a tractable, beam-limited exploration.

\section{Methodology}

The methodology is presented from four perspectives: (1) The definition of a domain-specific tokenization strategy that maps physical laminate properties to discrete sequence tokens, (2) The architecture of the SeqGPT agent, (3) The dataset generation and training details, and (4) The neuro-symbolic decoding protocol for multi-panel generation.
\subsection{AI Modeling and Tokenization Strategy}

As introduced in Section 2.1, the generative model is conditioned on the continuous target vector $\boldsymbol{\xi}^*$ and the target laminate thickness $N$. To integrate these physical targets into the neural architecture, $\boldsymbol{\xi}^*$ and $N$ are concatenated and directly mapped into the model's latent space via a linear projection layer.

\textbf{The Symmetry constraint (MC1)} optimizes the generation process: the neural agent predicts only the first half of the sequence ($n = \lceil N/2 \rceil$ plies). The full laminate is then reconstructed by exact mirroring, with the final generated token acting as the central axis for odd number of plies.

\textbf{Next-token prediction.} Standard angles $\{0^\circ, 90^\circ, \pm 45^\circ\}$ are mapped to integer tokens $\{1, 2, 3, 4\}$, alongside \texttt{[PAD]} ($0$) and \texttt{[EOS]} ($5$) tokens for training. During inference, since the target thickness $N$ is provided in the prompt, generation stops deterministically once the required $n$ plies are produced.

\textbf{Initialization Prompt.} To ensure outer surface protection (Damage Tolerance), we enforce a fixed boundary condition where all generated laminates must begin with a $[45^\circ, -45^\circ]$ block, mapped to tokens $[3, 4]$. In our Decoder-Only architecture, the autoregressive generation is triggered by an initial context acting as a prompt and denoted as \texttt{[BOS]}. This initial state, \texttt{[BOS]}, encapsulates the physical conditioning vector $[PHYS] = (\boldsymbol{\xi}^*, N)$ alongside these fixed boundary tokens. This initializes the sequence as \texttt{[BOS]} $= [PHYS, 3, 4]$, ensuring that every generated layup is physically conditioned from the first step and satisfies the mandatory surface requirements by construction.

\subsection{Architecture and Input Conditioning}
\label{sec:architecture}

\begin{figure}[httbp]
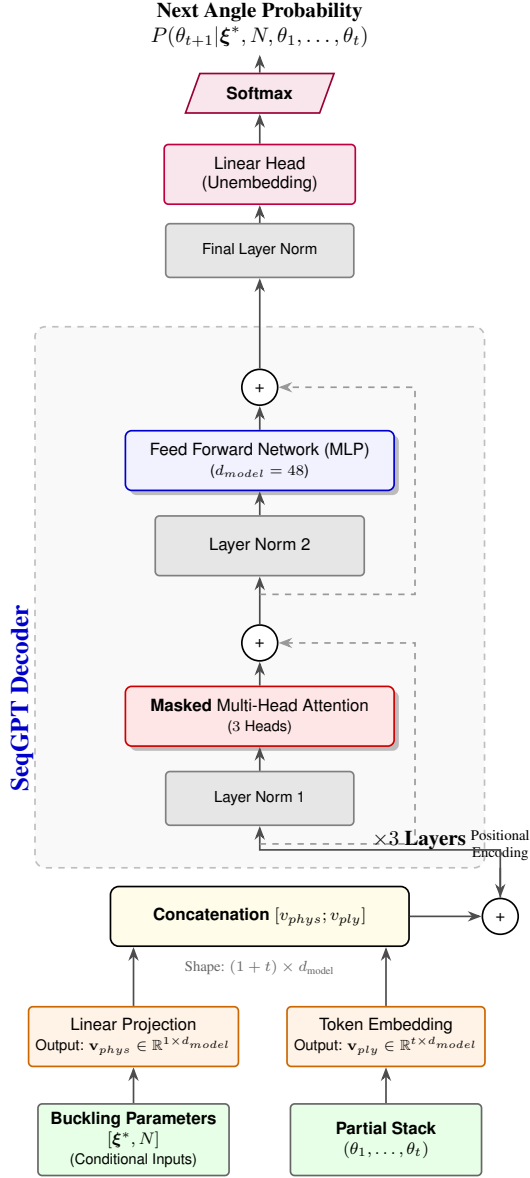

    \centering
    \includestandalone[width=0.4\textwidth]{figures/fig_seqgpt}
    \caption{Architecture of the SeqGPT agent. The model is a lightweight Decoder-Only Transformer.}
    \label{fig:seqgpt}
\end{figure}

We employ a conditional Transformer architecture based on the \textbf{minGPT} framework \cite{karpathy2020mingpt} (see Figure~\ref{fig:seqgpt}). The model is a lightweight Transformer Decoder composed of \textbf{3 layers} with \textbf{3 attention heads}. The embedding dimension is set to $d_{model} = 48$, resulting in approximately 83k trainable parameters. Following the standard minGPT architecture, each transformer block incorporates Layer Normalization, Masked Multi-Head Attention, and a Multilayer Perceptron (MLP) featuring a GELU non-linear activation function. Dropout layers are also included to act as a regularization technique and prevent overfitting.

We have two different input types: conditional input parameters $[PHYS] = (\boldsymbol{\xi}^*, N)$ and discrete ply tokens $\{1,2,3,4\}$. To mix them, we project them into a shared space of dimension $d_{model}$.

Because Transformers require fixed-length input vectors while optimal laminates exhibit variable thicknesses, we standardize the sequence capacity based on our dataset's maximum thickness, $N_{\max} = 41$. Consequently, we define a maximum context window of $n_{\max} = \lceil N_{\max}/2 \rceil = 21$. During inference, SeqGPT autoregressively generates the half-sequence until $n$ ply tokens have been produced. Any remaining sequence length up to $n_{\max}$ is subsequently filled with \texttt{[PAD]} tokens.

\textbf{Feature Fusion and Positional Encoding}.We use two projection layers to enter the common latent space:
\begin{itemize}
    \item The \textbf{conditional input} vector of physical targets is processed by a Linear Layer. This projects floating-point values into vectors $v_{phys} \in \mathbb{R}^{d_{model}}$.
    \item The \textbf{discrete tokens} pass through a standard trainable Embedding Layer. This maps each integer to a dense vector $v_{ply} \in \mathbb{R}^{d_{model}}$.
\end{itemize}

We concatenate these vectors to form the input sequence, with the physical parameters prepended as a conditioning prompt for generation. Because transformers treat tokens as unordered, this architecture adds learned positional embeddings to the sequence vectors. This allows the model to capture the exact positional context of each token within the stack.

\textbf{Attention Mechanism and Physical Alignment}. During the self-attention calculation, every new token generated at step $t$ attends back to all previous positions $\theta_{<t}$. Crucially, it also attends to the initial positions corresponding to $v_{phys}$.
The attention weights implicitly learn to correlate specific ply orientations to match the physical target.

\textbf{Training Objective and Masking}. The model operates autoregressively: at each step it predicts the probability distribution of the next token conditioned on the sequence history and the physical context. The network is trained by minimizing the Cross-Entropy loss. However, the conditional input token \texttt{[PHYS]} and the padding token \texttt{[0]} must be excluded from the optimization.
To achieve this, a binary mask $m_t$ is introduced such that $m_t = 0$ for positions corresponding to physical parameters or padding tokens, and $m_t = 1$ for valid ply tokens.
For a sequence with maximum length $n_{\max}$, the masked Cross-Entropy loss over the vocabulary space $V$ is defined as
\begin{equation}
\mathcal{L} = - \sum_{t=1}^{n_{\max}} m_t \sum_{j=1}^{|V|} \hat{y}_{j,t} \log(y_{j,t}),
\end{equation}
where $|V|$ denotes the vocabulary size, $\hat{y}_{j,t} \in \{0,1\}$ is the one-hot encoded ground-truth token for class $j$ at position $t$, and 
\[
y_{j,t} = P(\theta_t = j \mid \theta_{<t}, \boldsymbol{\xi}^*, N)
\]
is the probability predicted by the softmax layer for token $j$ at step $t$, conditioned on the previous plies and the target physical parameters.
The objective optimized during training corresponds to the average of this loss over all tokens in a mini-batch.

\subsection{Dataset and Training Details}
The model was evaluated after being trained on a synthetic dataset of 20,000 unique, valid stacking sequences. The dataset was generated via an Acceptance-Rejection Monte Carlo sampling strategy: an extensive pool of over $10^6$ randomly generated discrete layups was evaluated, and only those strictly satisfying the local manufacturing rules (MC1-MC5) were retained. For each valid sequence, the corresponding Nemeth parameters ($\alpha, \beta, \gamma, \delta$) were analytically computed using the Classical Laminate Theory (CLT) based on IMA/M21E material properties ($E_1=154$ GPa, $E_2=8.5$ GPa, $G_{12}=4.2$ GPa, $\nu_{12}=0.35$).

The SeqGPT model uses a 'gpt-nano' architecture configuration with a block size of 50. It was trained using the AdamW optimizer with a weight decay of $0.1$, $\beta$ parameters of $(0.9, 0.95)$, and a constant learning rate of $3 \times 10^{-4}$. To prevent gradient explosion, gradients were clipped at a maximum norm of $1.0$. We used a batch size of $64$ and trained the model for $30,000$ iterations. Cross-entropy loss was calculated exclusively on the valid ply tokens using the masking strategy described in Eq. 2.
\subsection{Multi-Panel Generation via Constrained Decoding}
\label{sec:blending}
SeqGPT efficiently generates isolated laminates that satisfy all local rules (MC1--MC5). However, multi-panel blending (MC0) requires deriving a child sequence $S_c$ from a parent $S_p$. While SeqGPT's autoregressive nature intrinsically respects Markovian constraints like contiguity (MC2) and disorientation (MC4), it lacks visibility over the parent's available ply inventory. Consequently, locally optimal predictions can render the sequence infeasible later—for instance, dropping the parent's only $0^\circ$ ply irreversibly violates the child's proportion constraint (MC5).

To prevent such conflicts and guarantee that the neural predictions align with both the blending topology (MC0) and the cumulative manufacturing rules (MC3, MC5), a child laminate is considered \emph{blendable} with respect to its parent if and only if it satisfies the set of \textbf{hard blending rules} defined in Table~\ref{tab:blending_constraints}. These rules (C1--C3) act as explicit symbolic masks within the decoding loop. Multi-panel generation therefore amounts to searching, in the discrete space of orientation sequences, for paths that remain valid under these combined constraints.

This induces a constrained tree search where each node corresponds to a partial stacking sequence and each edge to the addition of one ply. At each depth $t$, if a candidate ply violates any of the hard constraints (either by breaking the parent's topology or exhausting a critical orientation), the entire branch is pruned. 

\begin{figure}[htbp]
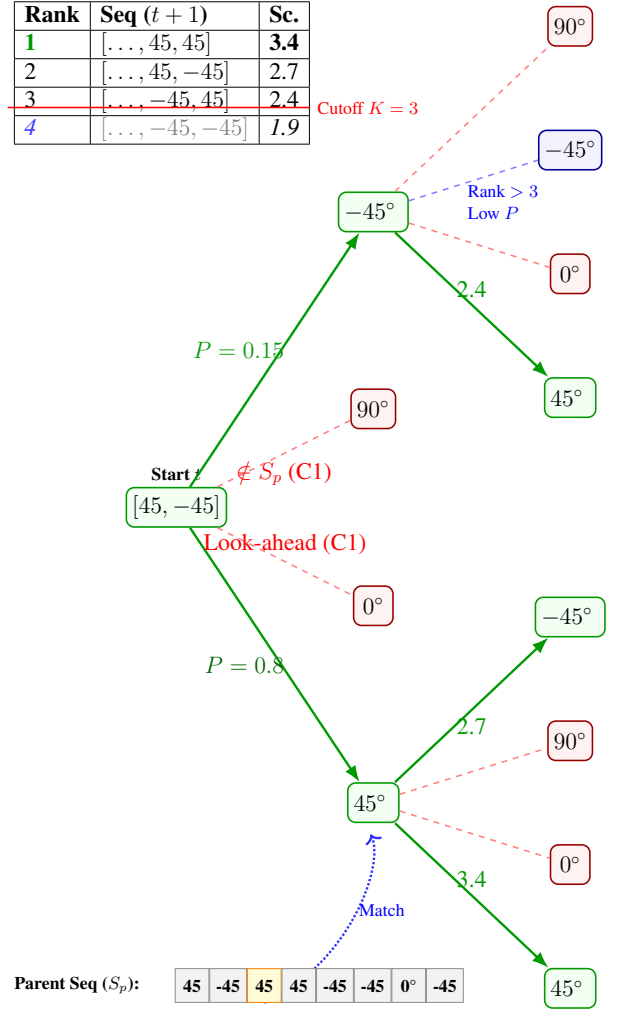

    \centering
    \includestandalone[width=0.45\textwidth]{figures/fig_tree}
    \caption{Neuro-Symbolic Constrained Beam Search for laminate blending. Candidate plies proposed by SeqGPT are filtered by hard blending rules with respect to the parent sequence $S_p$, then ranked using a probability–penalty score. Only the top-$K$ valid branches are retained at each step.}
    \label{fig:neuro_search}
\end{figure}

Figure~\ref{fig:neuro_search} illustrates the constrained beam search procedure with a beam width of $K=3$. At each step, the decoder retains the top-K ranked valid branches, transforming the exponential search into a tractable beam-limited exploration guided by SeqGPT.

SeqGPT acts as a probabilistic guide to rank the remaining valid candidates. Given a partial sequence $\theta_{\leq t}$, the cumulative score of a branch extended with a candidate ply $\theta_{t+1}$ is:
\begin{equation}
\text{Score}(\theta_{\leq t+1}) = \text{Score}(\theta_{\leq t}) + \log P(\theta_{t+1} \mid \theta_{\leq t}, \boldsymbol{\xi}^*, N),
\end{equation}
where the conditional log-probability term is directly approximated by the neural predictions of the SeqGPT agent. This scoring function favors mechanically plausible continuations conditioned on the optimal continuous parameters $\boldsymbol{\xi}^*$ and the target thickness $N$.

\begin{table}[htbp]
\centering 
\footnotesize
\caption{Decoding Constraints for Blending. Hard constraints (C) strictly prune the search space to ensure manufacturing feasibility.} 
\label{tab:blending_constraints} 
\renewcommand{\arraystretch}{1.2} 
\begin{tabular}{|l|p{2.2cm}|p{3.8cm}|} 
\hline \textbf{ID} & \textbf{Constraint} & \textbf{Description / Condition} \\ \hline 
\textbf{C1} & Sequence Heritage & \textbf{Drop:} Inherit plies from $S_p$ sequentially ($S_c \subset S_p$). The total number of skipped parent plies must not exceed $N_p - N_c$. \newline \textbf{Insertion:} Inherit all plies from $S_p$ sequentially ($S_p \subset S_c$). The total number of newly generated plies must not exceed $N_c - N_p$. \\ \hline 
\textbf{C2} & Balance & $S_p$ must provide counter-angles to offset current $S_c$ imbalance. \\ \hline 
\textbf{C3} & Proportion & Sufficient angles of each orientation remain to satisfy the ply proportion constraint. \\ \hline 
\end{tabular} 
\end{table}

\paragraph{Soft constraints and Fallback Strategy.}
Beyond hard pruning (C1--C3), additional manufacturing heuristics can be incorporated as soft penalties that are subtracted from the log-probability score. These penalties bias the beam ranking without affecting absolute feasibility. For instance, during a "drop" mode, a continuity penalty favors inheriting contiguous plies from the parent rather than creating large structural gaps.

Furthermore, to guarantee robustness, a \textbf{Conservative Fallback Strategy} is introduced. If the constrained beam search fails to produce a valid sequence for a target thickness $N_{\text{target}}$ (often due to overly restrictive constraints), the target thickness is iteratively increased. In the worst case, the parent stacking sequence is fully copied ($N_{\text{target}} = N_{\text{parent}}$). This ensures that a manufacturable solution is always returned.

\section{Numerical Results and Discussion}
This section presents the evaluation of the proposed framework, moving from intrinsic model assessment to industrial validation. Section 4.1 first characterizes the generative performance independently of the beam search solver, evaluating the model's capacity to strictly respect manufacturing constraints, accurately target buckling parameters, and provide interpretability through self-attention mechanisms. Section 4.2 then validates the full neuro-symbolic framework on the standard 18-panel horseshoe benchmark. Finally, Section 4.3 discusses these results by comparison to the baseline in terms of mechanical margins and computational efficiency.

\subsection{Single Panel Generation with SeqGPT}

\subsubsection{Diversity--Validity Trade-off under Top-k Sampling}
We evaluated the generative behavior of SeqGPT independently of the deterministic beam-search solver by sampling 25 sequences for each of 200 buckling targets. Diversity was controlled through \textit{Top-$k$} sampling, with $k=1$ corresponding to greedy decoding and $k>1$ introducing stochastic exploration. No generated sequence appeared in the training set, indicating that the model learns physical correlations rather than memorizing configurations.

\begin{figure}[htbp]
    \centering
    \includegraphics[width=1\columnwidth]{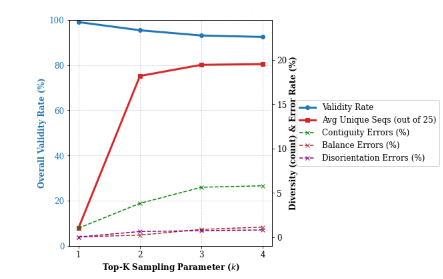}
    \caption{Evolution of the sequence validity rate (left axis) and the breakdown of specific constraint violations (right axis) as a function of the top-$K$ sampling parameter.}
    \label{fig:topk_error_breakdown}
\end{figure}

Figure~\ref{fig:topk_error_breakdown} shows that while the model captures most industrial rules, increasing $k$ degrades validity. Given the small vocabulary size ($|V|=4$), setting $k=4$ effectively removes any pruning of the output distribution. 
As a result, the sampler is forced to consider all possible ply orientations, including low-probability but physically invalid transitions (e.g., $0^\circ \rightarrow 90^\circ$), which leads to occasional violations of the disorientation rule.

A moderate setting ($k=2$) provides a trade-off between diversity and feasibility, but strict manufacturing compliance without external repair requires greedy decoding ($k=1$).

\subsubsection{Accuracy of Mechanical Target Conditioning}

An effective inverse design framework must accurately recover the continuous physical properties prescribed by the macroscopic optimization phase.
We evaluated the constrained model capacity to target specific parameters 
\begin{figure}[htbp]
    \centering
    \includegraphics[width=\columnwidth]{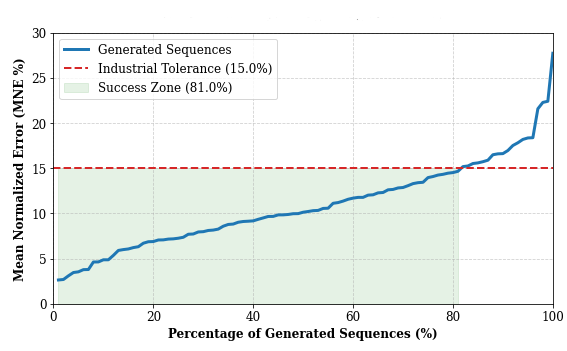}
    \caption{Cumulative distribution of the Mean Normalized Error (MNE) across 1000 generated sequences. The dashed line defines the 15.0\% industrial tolerance threshold.}
    \label{fig:success_rate}
\end{figure}
($\alpha, \beta, \gamma, \delta$) on 1000 unseen target vectors using greedy decoding ($k=1$).
Exact matching of the continuous targets is often impossible due to the discrete nature of ply angles ($\{0^\circ, \pm 45^\circ, 90^\circ\}$) and the cubic dependence of flexural stiffness. Consequently, targeting performance is evaluated using the Mean Normalized Error (MNE). Figure~\ref{fig:success_rate} displays the cumulative distribution of the targeting error. Using a 15\% MNE industrial tolerance threshold, the framework achieves an overall success rate of \textbf{81.0\%}.

These controlled error rates confirm that the latent conditioning mechanism guides the generative process toward the required mechanical regions. SeqGPT then acts as an efficient heuristic locator that guides the beam search presented in the previous section. 

\subsubsection{Model Interpretability: Self-Attention Analysis}
To understand the model's decision-making process, we analyzed the multi-head self-attention weights from the final transformer layer. These weights represent the network's focus on previously generated plies and conditioning targets during autoregressive decoding.

\begin{figure}[htbp]
    \centering
    \includegraphics[width=0.5\textwidth]{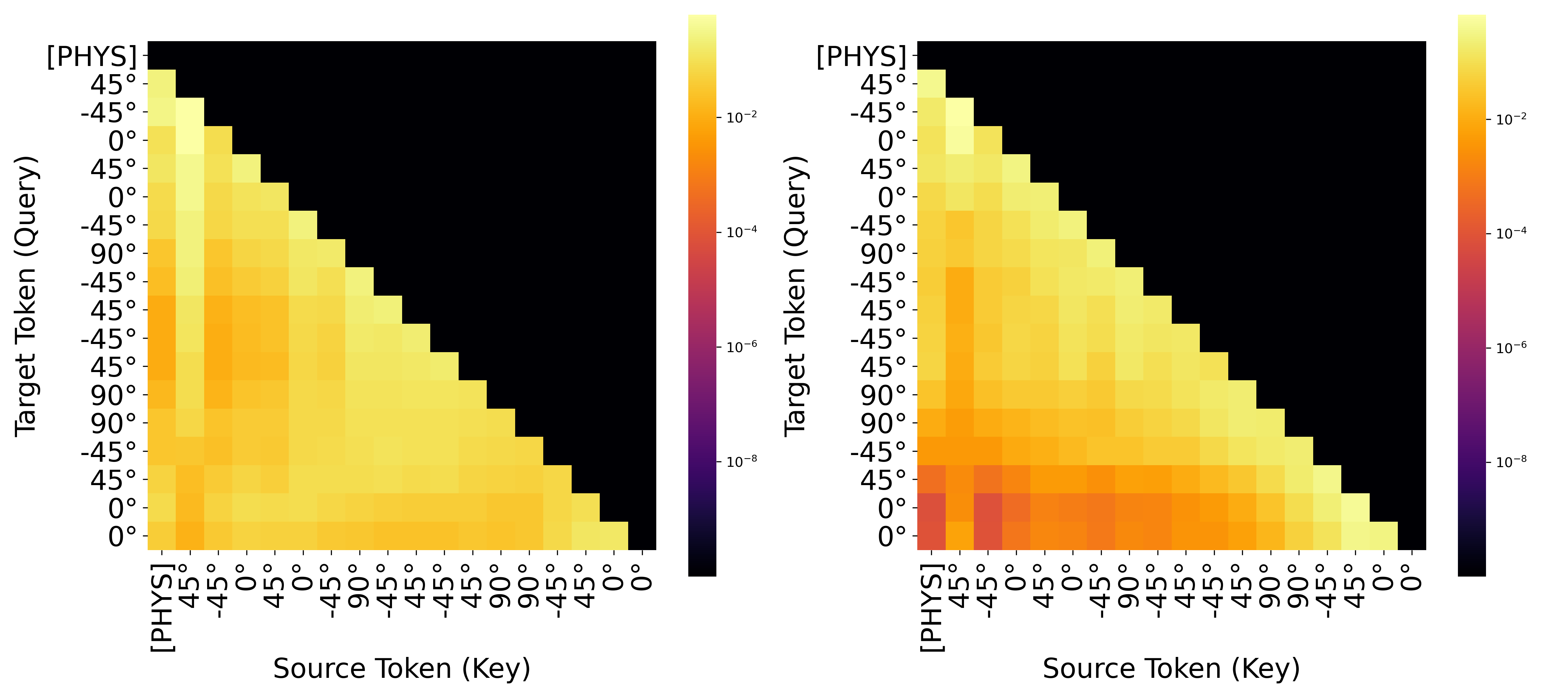}
    \caption{Self-attention weight distributions from the final layer of the SeqGPT model. (a) Head 1: Global context. (b) Head 2: Local Markovian constraints.}
    \label{fig:attention_heads}
\end{figure}

The visualizations in Fig. \ref{fig:attention_heads} suggest that the attention heads might capture different aspects of the sequence. As expected from causal masking, all matrices exhibit a lower-triangular structure. Head 2 (Fig. \ref{fig:attention_heads}b) shows a strong concentration of attention weights along the immediate sub-diagonal. This suggests a primary focus on the directly preceding ply, which is consistent with the model learning local Markovian constraints (such as contiguity or maximum disorientation). In contrast, Head 1 (Fig. \ref{fig:attention_heads}a) captures longer-range dependencies across the entire sequence history, which could indicate an aggregation of global context, potentially to monitor cumulative properties like the overall $\pm 45^\circ$ balance.

\textbf{Physical Alignment.} The first column of both matrices shows how the model attends to the initial conditioning token ([PHYS]). Head 1 maintains persistent attention on this token, ensuring continuous grounding in the physical targets. Conversely, Head 2's attention to [PHYS] progressively decays as the sequence lengthens. Towards the end of the generation, local ply-to-ply constraints (the sub-diagonal) become so dominant that the model focuses almost exclusively on the preceding ply and ignores both the distant sequence history and the initial physical token. The third attention head is omitted due to redundancy.

\subsection{Multi-Panel Generation: 18-Panel Horseshoe Benchmark}
\subsubsection{Test Case}
\begin{figure}[htbp]
    \centering
    \includegraphics[width=\columnwidth]{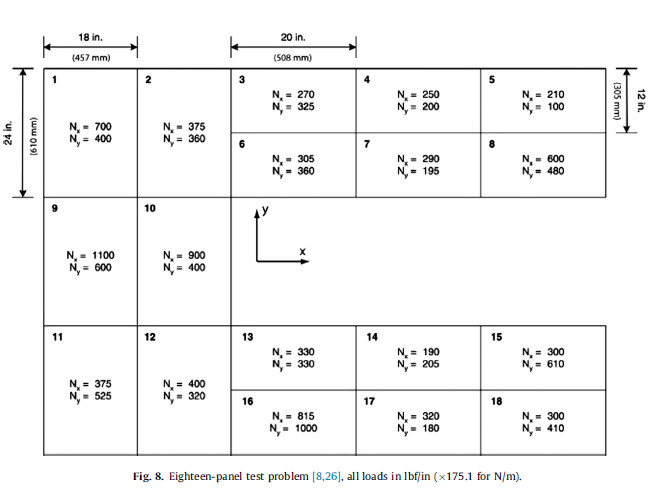}
    \caption{Eighteen-panel test problem benchmark layout \cite{irisarri2014}. All local loadings $N_x$ and $N_y$ are in lbf/in.}
    \label{fig:horseshoe_benchmark}
\end{figure}

The performance of the proposed neuro-symbolic approach is evaluated on the reference 18-panel horseshoe benchmark ($r = 18$) \cite{irisarri2014}. As shown in Figure~\ref{fig:horseshoe_benchmark}, this problem consists of interconnected panels with fixed biaxial loadings and dimensions. The structural material is Graphite/Epoxy IM7/8552 ($E_{1}=141$~GPa, $E_{2}=9.03$~GPa, $G_{12}=4.27$~GPa, $\nu_{12}=0.32$, ply thickness $0.191$~mm). Panel thickness is bounded between $n_{min}=14$ and $n_{max}=48$ plies.
While the original benchmark by Irisarri et al. \cite{irisarri2014} explores an extended orientation set (steps of 15$^\circ$), we restrict our design space to the industrial standard set $\{0^\circ, \pm 45^\circ, 90^\circ\}$. This choice aligns with conventional manufacturing certifications and focuses the challenge on achieving the target nemeth parameters with a more constrained, discrete vocabulary.
The objective is to minimize total mass while ensuring a buckling Reserve Factor $RF > 1$. We retrieve optimal thickness distributions and buckling parameters directly from \cite{irisarri2014} to use as physical targets for sequence generation.

\subsubsection{Propagation Map Definition}

To translate the local geometries and biaxial loadings defined in the test case (Fig.~\ref{fig:horseshoe_benchmark}) into manufacturable laminates, the multi-panel optimization process follows the standard bi-level strategy introduced in Section~\ref{sec:bi_level}. 

\textbf{The first step} of this framework---the continuous optimization phase---takes these geometric boundaries and internal loads ($N_x, N_y$) as inputs to minimize the total structural mass. This continuous relaxation yields the optimal target vector $\boldsymbol{\xi}^*$ alongside the target thickness $N_{\text{target}}$ for each individual panel $i$. This output can be formally defined as the set of continuous physical targets $\mathcal{T} = \{ (\boldsymbol{\xi}^*_i, N_{\text{target},i}) \}_{i=1}^{18}$. 

Because the core focus of our study is the discrete stacking sequence recovery, we bypass executing the continuous FEA ourselves. Instead, we directly extract this global optimum target set $\mathcal{T}$ from the reference solution provided by Irisarri et al. \cite{irisarri2014}. 

\begin{figure}[htbp]
    \centering
    \includegraphics[width=0.45\textwidth]{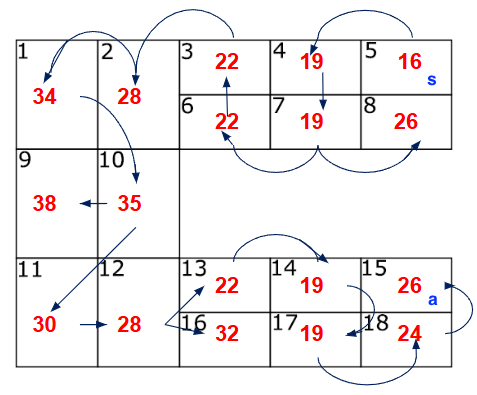} 
    \caption{Propagation map for the 18-panel benchmark. Arrows indicate the Search Propagation Direction (SPD). The integers denote the extracted target thicknesses $N_{\text{target},i}.$}
    \label{fig:spd_map}
\end{figure}

To solve the subsequent multi-panel recovery problem---\textbf{the second step} of the bi-level optimization framework---we manually prescribe a Propagation Map using a Search Propagation Direction (SPD) strategy \cite{scardaoni2022}. Figure~\ref{fig:spd_map} illustrates the topological layout for the 18-panel benchmark, where the integers denote the extracted target thicknesses $N_{\text{target},i}$. The map dictates the sequence generation order, propagating the design from designated start sequences to end sequences across adjacent panels. This definition effectively reduces the multi-panel blending problem into a sequential generation task, where topological continuity constraints are intrinsically handled by the neuro-symbolic beam search.

\subsubsection{Comparative Analysis: SeqGPT vs. Evolutionary Algorithms}

To assess the structural validity of the generated sequences, we compute the critical buckling load factor ($\lambda_{cr}$) for each panel. The resistance to buckling is governed by the panel's bending stiffness matrix $[D]$, which is directly derived from the stacking sequence and correlates with the lamination parameters ($\alpha, \beta, \gamma, \delta$) targeted by the model.

For a simply supported rectangular plate of dimensions $a \times b$ subjected to biaxial compression ($N_x, N_y$), the buckling factor for a mode $(m,n)$ is given by the analytical solution used in \cite{irisarri2014}:

\begin{equation}
    \resizebox{1.0\hsize}{!}{$
    \lambda_{(m,n)} = \frac{\pi^2 \left[ D_{11} \left(\frac{m}{a}\right)^4 + 2(D_{12} + 2D_{66}) \left(\frac{m}{a}\right)^2 \left(\frac{n}{b}\right)^2 + D_{22} \left(\frac{n}{b}\right)^4 \right]}{ \left(\frac{m}{a}\right)^2 N_x + \left(\frac{n}{b}\right)^2 N_y }
    $}.
    \label{eq:buckling}
\end{equation}

The Reserve Factor (RF) is defined as the minimum factor over all failure modes: $RF = \min_{m,n} \lambda_{(m,n)}$. A panel is considered safe if $RF > 1$. The values of $a$,$b$,$N_{x}$, and $N_{y}$ are shown in Figure~\ref{fig:horseshoe_benchmark}.

In this study, stability is reported using the Buckling Margin:
\begin{equation}
    \text{Margin (\%)} = (RF - 1) \times 100
\end{equation}

Generated sequences are compared against "Solution 0" from Irisarri et al. \cite{irisarri2014}, which serves as the reference for target thicknesses ($N_{\text{target}}$) and buckling margins.

Table~\ref{tab:final_results} details the comparative results. Notably, for all 18 panels, SeqGPT generated fully blended stacking sequences matching the exact target thickness ($N_{\text{generated}} = N_{\text{target}}$). Because no sequence required an artificial thickness increment to satisfy the blending rules, \textbf{the conservative fallback strategy was never triggered}. This demonstrates the efficiency of the neuro-symbolic beam search in navigating a highly constrained design space on its first attempt, yielding structural buckling margins that are highly competitive with the reference Evolutionary Algorithm (EA), while avoiding any mass penalty.

\begin{table}[ht]
    \centering
    \footnotesize
    \setlength{\tabcolsep}{4pt} 
    \renewcommand{\arraystretch}{0.9} 
    \caption{Comparison of SeqGPT generated solutions vs. Evolutionary Algorithm (EA) (Irisarri et al. \cite{irisarri2014}).}
    \label{tab:final_results}
    \begin{tabular}{@{} c c cc cc @{}} 
        \toprule
        \textbf{Panel} & \textbf{Target} & \multicolumn{2}{c}{\textbf{SeqGPT (Proposed)}} & \multicolumn{2}{c}{\textbf{EA Reference}} \\
        \cmidrule(lr){3-4} \cmidrule(l){5-6}
        & $N_{\text{target}}$ & $N_{\text{generated}}$& Margin (\%) & Margin (\%) \\ 
        \midrule
        1 & 34 & 34 & 16.1 & 16.6 \\
        2 & 28 & 28 & 3.1  & 3.7  \\
        3 & 22 & 22 & 13.4 & 14.5 \\
        4 & 19 & 19 & 6.6  & 7.7  \\
        5 & 16 & 16 & 5.3  & 6.5  \\
        6 & 22 & 22 & 1.9  & 2.9  \\
        7 & 19 & 19 & 3.2  & 4.3  \\
        8 & 26 & 26 & 8.5  & 9.6  \\
        9 & 38 & 38 & 3.4  & 4.8  \\
        10 & 35 & 35 & 3.6 & 4.5  \\
        11 & 30 & 30 & 10.4 & 10.3 \\
        12 & 28 & 28 & 4.3  & 3.3  \\
        13 & 22 & 22 & 2.7  & 7.8  \\
        14 & 19 & 19 & 9.4  & 14.3 \\
        15 & 26 & 26 & 6.2  & 6.2  \\
        16 & 32 & 32 & 8.5  & 7.9  \\
        17 & 19 & 19 & 1.3  & 5.8  \\
        18 & 24 & 24 & 16.2 & 20.5 \\
        \bottomrule
    \end{tabular}
\end{table}

\subsection{Discussion: Design Space, Performance, and Computational Cost}
Comparing our generated sequences with the EA reference \cite{irisarri2014} highlights three key methodological differences:

\begin{enumerate}
    \item \textbf{Restricted Design Space:} The reference EA operates on a wide range of orientations ($\{0^\circ, \pm 15^\circ, \dots, 90^\circ\}$). In contrast, SeqGPT is strictly restricted to standard aerospace angles $\{0^\circ, \pm 45^\circ, 90^\circ\}$ to comply with industrial manufacturing standards. Despite this significant reduction in the design space, our approach yields competitive results that remain close to the baseline.

    \item \textbf{Computational Efficiency:} The stochastic nature of EAs typically requires thousands of evaluations (approx. one hour for the reference study~\cite{irisarri2014}). The proposed constructive approach dynamically prunes invalid sequences during the beam search, generating the entire set of blended sequences in a single inference pass. This reduces the total recovery time to under \textbf{4.5 seconds} (with a beam width of 10).
    
    \item \textbf{System Maintainability:} The explicit decoupling between the data-driven physical model (neural) and the manufacturing rules (symbolic) provides a major advantage for the system's lifecycle. While the symbolic decoder is primarily mandated to enforce the global blending constraint, it can readily be used to filter local rules as well. If industrial manufacturing constraints evolve, these symbolic filters can be updated directly, without requiring costly dataset regeneration or model retraining.
\end{enumerate}

\section*{Conclusion}

This paper introduced SeqGPT, a neuro-symbolic framework for the inverse design of blended composite structures under manufacturing constraints. By reformulating stacking sequence retrieval as a conditional sequence generation problem and combining a lightweight Transformer with constrained decoding, the approach eliminates the main computational bottleneck of classical bi-step composite optimization workflows.

The key contribution lies in the explicit decoupling between data-driven modeling of mechanical correlations and symbolic enforcement of blending rules, ensuring manufacturable solutions by construction. Results on the 18-panel horseshoe benchmark demonstrate buckling performances comparable to state-of-the-art evolutionary approaches, with recovery times reduced from hours to seconds.

Several directions for future work are identified to address current limitations and extend the framework. 
First, the current formulation aggregates all physical parameters into a single context embedding. Future developments will investigate more structured conditioning schemes based on multiple context tokens, enabling finer control of the self-attention mechanism and improving the alignment between continuous performance targets and discrete stacking sequence realizability.
Second, multi-panel composite structures are naturally represented as graphs. While the current Search Propagation Direction (SPD) is manually defined through template-based strategies, future work will explore systematic graph-based formulations of propagation rules. Such an approach would enable the automatic inference of propagation paths for complex topologies while preserving the guarantees provided by symbolic constrained decoding.

\section*{Acknowledgments}
This research was supported by Airbus Operations SAS. This work was carried out as part of the ModIA apprenticeship program (INSA Toulouse and Toulouse INP-ENSEEIHT).

\section*{Generative AI Usage}
In accordance with the European Commission's recommendations, the authors acknowledge the use of generative AI (Google Gemini) for English language revision and assistance with TikZ figure generation. The authors assume full responsibility for all scientific content, methodologies, and conclusions presented in this manuscript.
\bibliographystyle{plain}
\bibliography{biblio-ch-pfia}
\end{document}

%% file: figures/fig_workflow.tex
\begin{tikzpicture}[
    node distance=0.8cm,
    font=\sffamily\footnotesize,
    >=Stealth,
    process/.style={
        rectangle, 
        minimum width=5.0cm, 
        minimum height=0.8cm, 
        text centered, 
        draw=black, 
        thick,
        fill=white,
        drop shadow
    },
    decision/.style={
        diamond, 
        minimum width=2.0cm, 
        minimum height=0.8cm, 
        text centered, 
        draw=black, 
        fill=gray!10,
        aspect=2
    },
    input/.style={
        trapezium, 
        trapezium left angle=70, 
        trapezium right angle=110, 
        minimum width=1.2cm, 
        text centered, 
        draw=black, 
        fill=white,
        inner sep=4pt
    },
    constraint/.style={
        rectangle,
        minimum width=2.2cm,
        draw=black,
        dashed,
        font=\scriptsize
    },
    step1/.style={process, fill=cyan!10},
    step2/.style={process, fill=orange!10},
    arrow/.style={thick, ->}
]

    \node (geo)  [input] {Geometry};
    \node (load) [input, left=0.2cm of geo] {Loads};
    \node (mat)  [input, right=0.2cm of geo] {Mat. Props};

    \node (s1) [step1, below=0.8cm of geo] {
        \textbf{1) Continuous Opt.} (Gradient / LPs)
    };
    \node (c1) [constraint, right=0.4cm of s1] {Step-1 Constraints};

    \node (s2) [step2, below=1.0cm of s1] {
        \textbf{2) Discrete Retrieval} (GA)
    };
    \node (c2) [constraint, right=0.4cm of s2] {Step-2 Constraints};

    \node (dec) [decision, below=0.8cm of s2] {Check?};
    \node (out) [process, below=0.8cm of dec, fill=green!10, minimum width=2.5cm] {Final Design};

    \draw [arrow] (geo) -- (s1.north);
    \draw [arrow] (load.south) -- ($(s1.north)+(-1.5,0)$);
    \draw [arrow] (mat.south) -- ($(s1.north)+(1.5,0)$);

    \draw [arrow] (c1) -- (s1.east);
    \draw [arrow] (c2) -- (s2.east);

    \draw [arrow] (s1) -- (s2);
    \draw [arrow] (s2) -- (dec);

    \draw [arrow] (dec) -- node[right, font=\scriptsize] {YES} (out);
    
    \draw [arrow] (dec.west) -- node[above, font=\scriptsize, xshift=-0.5cm] {NO} ++(-2,0) |- (s2.west);

\end{tikzpicture}

%% file: figures/fig_blending_cross_section.tex
\begin{tikzpicture}[
    scale=1.0, 
    font=\sffamily\scriptsize, 
    >=Stealth,
    ply0/.style={fill=cyan!30, draw=black, thin},      
    ply45/.style={fill=orange!30, draw=black, thin},   
    plyM45/.style={fill=yellow!30, draw=black, thin},  
    plyDrop/.style={fill=green!30, draw=black, thin},   
    resin/.style={fill=gray!20, pattern=north east lines, pattern color=gray!50, draw=none}
]

    \draw[ply0] (0,0) rectangle (6.5, 0.3);
    \node at (0.75, 0.15) {$0^\circ$};
    \node at (2.9, 0.15) {$0^\circ$};
    \node at (5.4, 0.15) {$0^\circ$};

    \draw[plyDrop] (0,0.3) rectangle (1.5, 0.6);
    \node at (0.75, 0.45) {$90^\circ$};
    \draw[resin] (1.5,0.3) -- (1.5,0.6) -- (2.2,0.3) -- cycle;

    \draw[resin] (4.3,0.3) -- (3.6,0.3) -- (4.3,0.6) -- cycle;
    \draw[ply45] (4.3,0.3) rectangle (6.5, 0.6);
    \node at (5.4, 0.45) {$45^\circ$};

    \draw[ply45] (0,0.6) rectangle (1.5, 0.9);
    \draw[ply45] (1.5,0.6) -- (2.2,0.3) -- (2.2,0.6) -- (1.5,0.9) -- cycle;
    \draw[ply45] (2.2,0.3) rectangle (3.6, 0.6);
    \draw[ply45] (3.6,0.3) -- (4.3,0.6) -- (4.3,0.9) -- (3.6,0.6) -- cycle;
    \draw[ply45] (4.3,0.6) rectangle (6.5, 0.9);
    \node at (0.75, 0.75) {$45^\circ$};
    \node at (2.9, 0.45) {$45^\circ$};
    \node at (5.4, 0.75) {$45^\circ$};

    \draw[plyDrop] (0,0.9) rectangle (1.5, 1.2);
    \node at (0.75, 1.05) {$90^\circ$};
    \draw[resin] (1.5,0.9) -- (1.5,1.2) -- (2.2,0.6) -- cycle;
    \draw[resin] (4.3,0.9) -- (3.6,0.6) -- (4.3,1.2) -- cycle;
    \draw[plyM45] (4.3,0.9) rectangle (6.5, 1.2);
    \node at (5.4, 1.05) {$-45^\circ$};

    \draw[plyM45] (0,1.2) rectangle (1.5, 1.5);
    \draw[plyM45] (1.5,1.2) -- (2.2,0.6) -- (2.2,0.9) -- (1.5,1.5) -- cycle;
    \draw[plyM45] (2.2,0.6) rectangle (3.6, 0.9);
    \draw[plyM45] (3.6,0.6) -- (4.3,1.2) -- (4.3,1.5) -- (3.6,0.9) -- cycle;
    \draw[plyM45] (4.3,1.2) rectangle (6.5, 1.5);
    \node at (0.75, 1.35) {$-45^\circ$};
    \node at (2.9, 0.75) {$-45^\circ$};
    \node at (5.4, 1.35) {$-45^\circ$};

    \draw[ply0] (0,1.5) rectangle (1.5, 1.8);
    \draw[ply0] (1.5,1.5) -- (2.2,0.9) -- (2.2,1.2) -- (1.5,1.8) -- cycle;
    \draw[ply0] (2.2,0.9) rectangle (3.6, 1.2);
    \draw[ply0] (3.6,0.9) -- (4.3,1.5) -- (4.3,1.8) -- (3.6,1.2) -- cycle;
    \draw[ply0] (4.3,1.5) rectangle (6.5, 1.8);
    \node at (0.75, 1.65) {$0^\circ$};
    \node at (2.9, 1.05) {$0^\circ$};
    \node at (5.4, 1.65) {$0^\circ$};

    \draw[decoration={brace, mirror, raise=5pt}, decorate, thick] (0,0) -- (1.5,0) 
        node[midway, below=10pt] {\textbf{Panel} $i-1$};
    \draw[decoration={brace, mirror, raise=5pt}, decorate, thick] (2.2,0) -- (3.6,0) 
        node[midway, below=10pt] {\textbf{Panel} $i$};
    \draw[decoration={brace, mirror, raise=5pt}, decorate, thick] (4.3,0) -- (6.5,0) 
        node[midway, below=10pt] {\textbf{Panel} $i+1$};

    \draw[thick, green!60!black, ->] (1.85, 2.3) -- (1.85, 1.7);
    \node[align=center, font=\bfseries\tiny, text=green!60!black, fill=white, inner sep=1pt] at (1.85, 2.5) 
        {Valid ($S_i \subset S_{i-1}$)};

    \draw[thick, green!60!black, ->] (4.0, 2.3) -- (4.0, 1.7);
    \node[align=center, font=\bfseries\tiny, text=green!60!black, fill=white, inner sep=1pt] at (4.0, 2.5) 
        {Valid ($S_i \subset S_{i+1}$)};

\end{tikzpicture}

%% file: figures/figspd.tex
\begin{tikzpicture}[
    font=\sffamily,
    >={Stealth[scale=1.2]},
    gridline/.style={draw=black, thick},
    number/.style={font=\Large\bfseries},
    dropPath/.style={->, dashed, thick, draw=black},
    insertPath/.style={->, thick, draw=black},
    sourceLabel/.style={font=\bfseries\footnotesize, text=black},
    addedLabel/.style={font=\bfseries\footnotesize, text=black},
    captiontext/.style={font=\normalsize\bfseries},
    legendtext/.style={font=\small}
]

\def\hgap{3.8} 


\draw[gridline] (0,0) rectangle (3,3);
\draw[gridline] (1.5,0) -- (1.5,3);
\draw[gridline] (0,1.5) -- (3,1.5);

\node[number] (a11) at (0.75,2.25) {11};
\node[number] (a19) at (2.25,2.25) {19};
\node[number] (a31) at (0.75,0.75) {31};
\node[number] (a8)  at (2.25,0.75) {8};

\node[sourceLabel] at (2.75,0.25) {$s$};
\node[addedLabel] at (0.25,0.25) {$a$};

\draw[insertPath] (a8) -- (a11);
\draw[insertPath] (a11) -- (a19);
\draw[insertPath] (a19) -- (a31);

\node[captiontext] at (1.5,-0.7) {(a)};


\begin{scope}[xshift=\hgap cm]

\draw[gridline] (0,0) rectangle (3,3);
\draw[gridline] (1.5,0) -- (1.5,3);
\draw[gridline] (0,1.5) -- (3,1.5);

\node[number] (b11) at (0.75,2.25) {11};
\node[number] (b19) at (2.25,2.25) {19};
\node[number] (b31) at (0.75,0.75) {31};
\node[number] (b8)  at (2.25,0.75) {8};

\node[sourceLabel] at (0.25,0.25) {$s$};
\node[addedLabel] at (2.75,1.75) {$a$};
\node[addedLabel] at (2.75,0.25) {$a$};

\draw[dropPath] (b31) -- (b11);
\draw[insertPath] (b11) -- (b19);
\draw[dropPath]   (b11) -- (b8);

\node[captiontext] at (1.5,-0.7) {(b)};

\end{scope}


\begin{scope}[xshift=2*\hgap cm]

\draw[gridline] (0,0) rectangle (3,3);
\draw[gridline] (1.5,0) -- (1.5,3);
\draw[gridline] (0,1.5) -- (3,1.5);

\node[number] (c11) at (0.75,2.25) {11};
\node[number] (c19) at (2.25,2.25) {19};
\node[number] (c31) at (0.75,0.75) {31};
\node[number] (c8)  at (2.25,0.75) {8};

\node[sourceLabel] at (2.75,0.25) {$s$};
\node[addedLabel] at (0.25,2.75) {$a$};
\node[addedLabel] at (2.75,2.75) {$a$};

\draw[insertPath]   (c8) -- (c31);
\draw[dropPath] (c31) -- (c11);
\draw[insertPath] (c8) -- (c19);

\draw[very thick, dotted] (0,0) rectangle (3,3);

\node[captiontext] at (1.5,-0.7) {(c)};

\end{scope}
\end{tikzpicture}

%% file: figures/fig_seqgpt.tex
\begin{tikzpicture}[
    font=\sffamily\footnotesize,
    >=Stealth,
    node distance=0.8cm,
    input/.style={
        rectangle, 
        draw=black!60, 
        fill=green!10, 
        thick, 
        minimum width=3.2cm, 
        minimum height=1.2cm, 
        align=center,
        rounded corners=2pt
    },
    embedding/.style={
        rectangle, 
        draw=orange!80!black, 
        fill=orange!10, 
        thick, 
        minimum width=3.2cm, 
        minimum height=1cm, 
        align=center,
        rounded corners=2pt
    },
    operation/.style={
        circle, 
        draw=black, 
        thick, 
        fill=white, 
        minimum size=0.6cm, 
        inner sep=0pt
    },
    block/.style={
        rectangle, 
        draw=blue!80!black, 
        thick, 
        fill=blue!5, 
        minimum width=4.5cm, 
        minimum height=1cm, 
        align=center,
        rounded corners=3pt,
        drop shadow
    },
    layer_box/.style={
        rectangle, 
        draw=gray!50, 
        dashed, 
        thick, 
        fill=gray!5, 
        inner sep=10pt, 
        rounded corners=5pt
    },
    output/.style={
        trapezium, 
        trapezium left angle=70, 
        trapezium right angle=110, 
        draw=purple!80!black, 
        fill=purple!10, 
        thick, 
        minimum width=2.5cm, 
        align=center
    },
    arrow/.style={->, thick, color=black!70}
]

    \node[input] (phys_input) {\textbf{Buckling Parameters}\\$[\boldsymbol{\xi}^*, N]$\\ \scriptsize (Conditional Inputs)};
    \node[embedding, above=0.6cm of phys_input] (lin_proj) {Linear Projection\\ \scriptsize Output: $\mathbf{v}_{phys} \in \mathbb{R}^{1 \times d_{model}}$};
    \draw[arrow] (phys_input) -- (lin_proj);

    \node[input, right=1cm of phys_input] (seq_input) {\textbf{Partial Stack}\\$(\theta_1, \dots, \theta_{t})$};
    \node[embedding, above=0.6cm of seq_input] (tok_emb) {Token Embedding\\ \scriptsize Output: $\mathbf{v}_{ply} \in \mathbb{R}^{t \times d_{model}}$};
    \draw[arrow] (seq_input) -- (tok_emb);

    
    \node[rectangle, draw=black, thick, fill=yellow!10, minimum width=5cm, minimum height=1.0cm, rounded corners, above=1.5cm of $(lin_proj)!0.5!(tok_emb)$] (concat) {\textbf{Concatenation} $[v_{phys}; v_{ply}]$};
    \node[below=0.1cm of concat, font=\scriptsize, color=gray!80!black] {Shape: $(1+t) \times d_{\text{model}}$};
    
    \draw[arrow] (lin_proj) -- (concat.south -| lin_proj);
    \draw[arrow] (tok_emb) -- (concat.south -| tok_emb);
    
    \node[operation, right=1.2cm of concat] (plus_pos) {+};
    \node[above=0.5cm of plus_pos, align=center, font=\scriptsize] (pos_enc) {Positional\\Encoding};
    \draw[arrow] (pos_enc) -- (plus_pos);
    \draw[arrow] (concat) -- (plus_pos);

    
    \node[embedding, fill=gray!20, draw=gray, minimum width=3.5cm, scale=0.9, above=1cm of concat] (norm1) {Layer Norm 1};
    \node[block, fill=red!10, draw=red!80!black, above=0.4cm of norm1] (attn) {\textbf{Masked} Multi-Head Attention\\ \scriptsize ($3$ Heads)};
    \node[operation, above=0.4cm of attn] (add1) {+};

    \node[embedding, fill=gray!20, draw=gray, minimum width=3.5cm, scale=1.0, above=0.8cm of add1] (norm2) {Layer Norm 2};
    \node[block, above=0.4cm of norm2] (ffn) {Feed Forward Network (MLP)\\ \scriptsize ($d_{model} = 48$)};
    \node[operation, above=0.4cm of ffn] (add2) {+};

    \begin{scope}[on background layer]
        \node[layer_box, fit=(norm1) (attn) (add1) (norm2) (ffn) (add2), inner xsep=1.5cm, inner ysep=0.7cm] (decoder_layer) {};
    \end{scope}

    \draw[arrow] (plus_pos) |- ($(norm1.south) + (0, -0.4)$) -- (norm1);
    \draw[arrow] (norm1) -- (attn);
    \draw[arrow] (attn) -- (add1);
    \draw[arrow] (add1) -- (norm2);
    \draw[arrow] (norm2) -- (ffn);
    \draw[arrow] (ffn) -- (add2);
    
    
    \draw[->, dashed, thick, color=gray!80] 
        ($(norm1.south) + (0, -0.3)$) -- 
        ++(2.6, 0) |- 
        (add1.east);

    \draw[->, dashed, thick, color=gray!80] 
        ($(norm2.south) + (0, -0.3)$) -- 
        ++(2.6, 0) |- 
        (add2.east);

    \node[anchor=south east, font=\bfseries, color=black, xshift=-0.2cm, yshift=0.2cm] at (decoder_layer.south east) {$\times 3$ Layers};
    \node[left=0.2cm of decoder_layer, rotate=90, font=\bfseries\large, color=blue!80!black] {SeqGPT Decoder};

    \node[embedding, fill=gray!20, draw=gray, minimum width=3.5cm, scale=0.9, above=0.8cm of decoder_layer] (norm_f) {Final Layer Norm};
    \node[embedding, fill=purple!10, draw=purple, above=0.3cm of norm_f] (unembed) {Linear Head\\(Unembedding)};
    \node[output, above=0.5cm of unembed] (softmax) {\textbf{Softmax}};
    \node[above=0.3cm of softmax, font=\bfseries, align=center] (prob) {Next Angle Probability\\$P(\theta_{t+1} | \boldsymbol{\xi}^*, N, \theta_{1}, \dots, \theta_{t})$};

    \draw[arrow] (add2) -- (norm_f);
    \draw[arrow] (norm_f) -- (unembed);
    \draw[arrow] (unembed) -- (softmax);
    \draw[arrow] (softmax) -- (prob);

\end{tikzpicture}

%% file: figures/fig_tree.tex
\begin{tikzpicture}[
    font=\small,
    grow=right, 
    level 1/.style={sibling distance=35mm, level distance=35mm},
    level 2/.style={sibling distance=22mm, level distance=35mm},
    edge from parent/.style={draw, thick, -Latex, gray!60},
    valid/.style={
        rectangle, 
        draw=green!60!black, 
        thick, 
        rounded corners, 
        fill=green!5, 
        minimum size=7mm, 
        inner sep=3pt,
        align=center,
        font=\sffamily\large\bfseries 
    },
    invalid/.style={
        rectangle, 
        draw=red!60!black, 
        thick, 
        rounded corners, 
        fill=red!5, 
        minimum size=7mm, 
        inner sep=3pt,
        font=\sffamily\large\bfseries 
    },
    valid_pruned/.style={ 
        rectangle, 
        draw=blue!60!black, 
        thick, 
        rounded corners, 
        fill=blue!5, 
        minimum size=7mm, 
        inner sep=3pt,
        font=\sffamily\large\bfseries 
    },
    accepted/.style={edge from parent/.style={draw, very thick, green!60!black, -Latex}},
    pruned/.style={edge from parent/.style={draw, thick, red!50, dashed, -}},
    parent_box/.style={
        draw=gray!90, 
        fill=gray!10, 
        minimum size=6mm, 
        font=\bfseries\small
    }
]

    \node[anchor=west] (label_p) at (-3, -8.5) {\textbf{Parent Seq ($S_p$):}};
    \node[parent_box, right=0.5cm of label_p] (p0) {45};
    \node[parent_box, right=0cm of p0] (p1) {-45};
    \node[parent_box, right=0cm of p1, fill=yellow!20, draw=orange] (p2) {45}; 
    \node[parent_box, right=0cm of p2] (p3) {45};
    \node[parent_box, right=0cm of p3] (p4) {-45};
    \node[parent_box, right=0cm of p4] (p5) {-45};
    \node[parent_box, right=0cm of p5] (p6) {0°};
    \node[parent_box, right=0cm of p6] (p7) {-45};

    \node[valid, label=above:{\textbf{Start $t$}}] (root) at (0,0) {$[45, -45]$}
        child[accepted] {
            node[valid] (n1) {$45^\circ$}
            child[accepted] { node[valid] {$45^\circ$} edge from parent node[above, font=\large] {3.4} }
            child[pruned] { node[invalid] {$0^\circ$} }
            child[pruned] { node[invalid] {$90^\circ$} }
            child[accepted] { node[valid] {$-45^\circ$} edge from parent node[below, font=\large] {2.7} }
            edge from parent node[above left, font=\large, color=green!40!black, pos=0.6] {$P=0.8$}
        }
        child[pruned] {
            node[invalid] (n2) {$0^\circ$}
            edge from parent node[above, font=\large, text=red] {\text{Look-ahead} (C1)}
        }
        child[pruned] {
            node[invalid] (n_new_90) {$90^\circ$}
            edge from parent node[below, font=\large, text=red] {$\notin S_p$ (C1)}
        }
        child[accepted] {
            node[valid] (n3) {$-45^\circ$}
            child[accepted] { 
                node[valid] {$45^\circ$} 
                edge from parent node[above, font=\large] {2.4}
            }
            child[pruned] { 
                node[invalid] {$0^\circ$} 
                edge from parent node[above, font=\large, text=red] {}
            }
            child { 
                node[valid_pruned] {$-45^\circ$} 
                edge from parent[draw, thick, blue!50, dashed, -]
                node[right, font=\small, text=blue, align=left, pos=0.4, yshift=-9pt] {Rank $>3$\\Low $P$}
            }
            child[pruned] { 
                node[invalid] {$90^\circ$} 
                edge from parent node[below, font=\large, text=red] {}
            }
            edge from parent node[below left, font=\large, pos=0.6] {$P=0.15$}
        };

    \node[anchor=north west, inner sep=0] (table) at (-3, 9) {
        \large
        \setlength{\tabcolsep}{5pt}
        \begin{tabular}{|l|l|l|}
            \hline
            \textbf{Rank} & \textbf{Seq ($t+1$)} & \textbf{Sc.} \\ \hline
            \textcolor{green!60!black}{\textbf{1}} & $[\dots, 45, 45]$ & \textbf{3.4} \\ \hline
            2 & $[\dots, 45, -45]$ & 2.7 \\ \hline
            3 & $[\dots, -45, 45]$ & 2.4 \\ \hline
            \textcolor{blue!80}{\textit{4}} & \textcolor{gray}{$[\dots, -45, -45]$} & \textit{1.9} \\ \hline
        \end{tabular}
    };
    
    \draw[red, thick] 
        ($(table.south west)!0.26!(table.north west)$) -- 
        ($(table.south east)!0.26!(table.north east)$) 
        node[right, font=\small, text=red] {Cutoff $K=3$};

    \begin{scope}[on background layer]
    \draw[->, densely dotted, very thick, blue!80, bend right=45, shorten >=29pt] 
        (p2.south) to node[midway, right, font=\small, text=blue] {Match} (n1.north west);
    \end{scope}
\end{tikzpicture}